\pdfoutput=1

\documentclass[11pt]{article}

\usepackage[]{acl}

\usepackage{times}
\usepackage{latexsym}
\usepackage{tabularx,booktabs}
\usepackage{rotating}
\usepackage{graphicx}

\usepackage{caption}
\usepackage{subcaption}

\usepackage[T1]{fontenc}

\usepackage[utf8]{inputenc}

\usepackage{microtype}

\usepackage{graphicx}
\usepackage{booktabs}
\usepackage{longtable}
\usepackage{tabu}
\usepackage{listings}
\usepackage{hyperref}
\usepackage{tcolorbox}
\usepackage{ctable}

\title{Exploring the Potential of Machine Translation for Generating Named Entity Datasets: A Case Study between Persian and English}

\author{
  Amir Sartipi \\ University of Isfahan\\ \texttt{amirsartipi.msc@eng.ui.ac.ir} \\ \And
  Afsaneh Fatemi \\ University of Isfahan\\ \texttt{a\_fatemi@eng.ui.ac.ir}
}
\begin{document}
\maketitle
\begin{abstract}
This study focuses on the generation of Persian named entity datasets through the application of machine translation on English datasets. The generated datasets were evaluated by experimenting with one monolingual and one multilingual transformer model. Notably, the CoNLL 2003 dataset has achieved the highest F1 score of 85.11\%. In contrast, the WNUT 2017 dataset yielded the lowest F1 score of 40.02\%. The results of this study highlight the potential of machine translation in creating high-quality named entity recognition datasets for low-resource languages like Persian. The study compares the performance of these generated datasets with English named entity recognition systems and provides insights into the effectiveness of machine translation for this task. Additionally, this approach could be used to augment data in low-resource language or create noisy data to make named entity systems more robust and improve them.


\end{abstract}

\section{Introduction}

\begin{figure}[tp]
        \centering
        \includegraphics[width= 0.95 \linewidth]{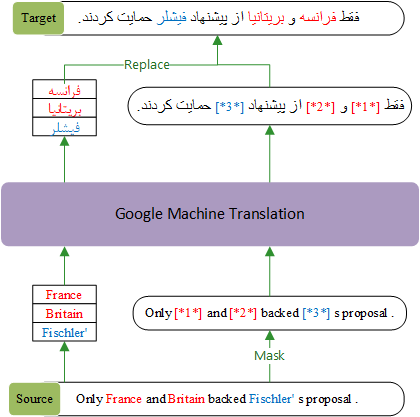}
        \caption{The process of translating an English sentence to Persian in the proposed approach}
        \label{fig:ch1-arch}
\end{figure}

Named Entity Recognition (NER) is a critical task in Natural Language Processing (NLP), with a wide range of applications including information extraction, sentiment analysis, and question-answering systems. However, the performance of NER systems is often hindered in low-resource languages, where there is a lack of annotated training data. Machine translation has shown promise in addressing this issue by providing a way to generate high-quality training datasets for low-resource languages. The process of translation for a sentence from English to Persian is shown in Figure \ref{fig:ch1-arch}.

The purpose of this paper can be summarized as follows:
\begin{enumerate}
    \item Generating high-quality Persian named entity recognition datasets using machine translation on English datasets.
    \item Evaluating the generated datasets using both monolingual and multilingual transformer models.
    \item Comparing the performance of the generated datasets with English counterpart named entity recognition systems in the Persian language and providing insights into the effectiveness of machine translation for this task.
\end{enumerate}

The article is structured as follows. In section \ref{sec:rw}, we provide a review of the most popular named entity recognition datasets in English and similar research that leverages machine translation to create datasets for low-resource languages. Additionally, we describe the existing named entity recognition datasets in the Persian language. In section \ref{sec:data}, we explain our methodology for generating the Persian named entity datasets using machine translation and highlight some key insights about the generated datasets. In section \ref{sec:expr}, we evaluate the performance of the generated datasets using transformer models, with both monolingual and multilingual model. In section \ref{sec:disc} we analyze the results of each dataset in detail. Finally, in section \ref{sec:conc}, we reach reasonable conclusions about our experiments, highlighting the potential of machine translation for named entity recognition in low-resource languages and the future directions for research in this area.

\begin{figure*}[htp]
        \centering
        \includegraphics[width= 0.95 \linewidth]{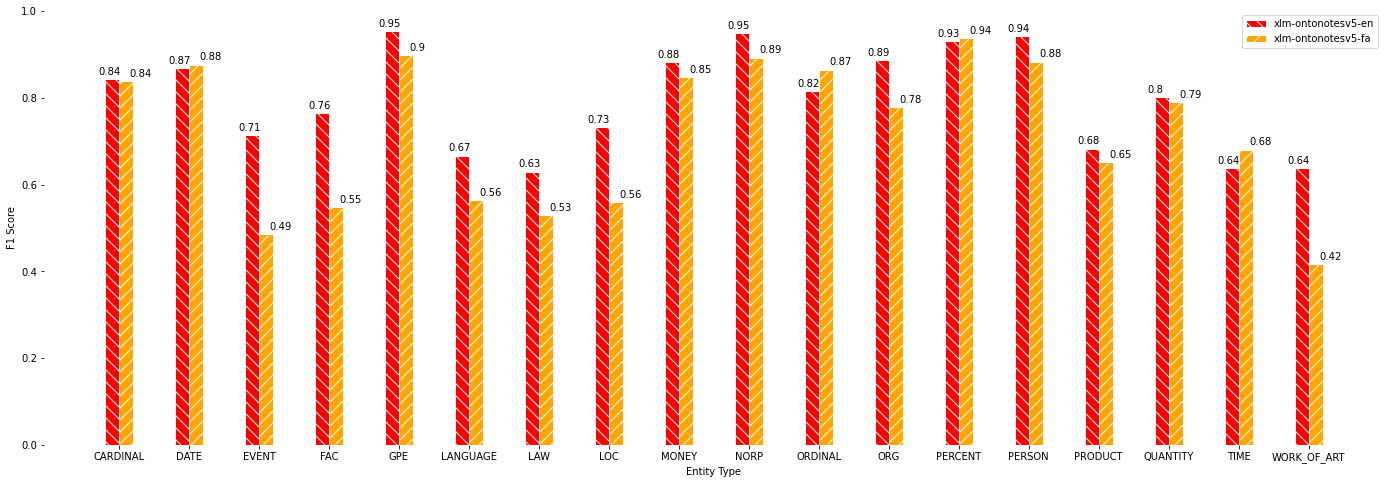}
        \caption{OntoNotes 5.0 F1 results on the English and translated data separated by entity types}
        \label{fig:ontov5}
\end{figure*}

\section{Related Work}
\label{sec:rw}
There exist several multilingual datasets that include Persian as a part of their data, and are distinguished by their coarse-grained \citep{multiconer-data} and fine-grained \citep{multiconer2-data} entities. However, established benchmarks such as CoNLL 2003 \citep{conll2033} and WNUT 2017 \cite{derczynski-etal-2017-results} do not include the Persian language in their respective corpora. In the following paragraphs, we elucidate the main characteristics of two Persian and four English datasets, along with some existing cross-lingual approaches.

\paragraph{Persian NER Datasets}

A corpus in Persian called "ArmanPerosNERCorpus" has been created, which includes 7,682 Persian sentences and 250,015 tokens. The dataset was manually annotated and categorized into six different classes of named entities, such as person, organization, location, facility, product, and event. The dataset has been made available in three folds for use as training and test sets \citep{poostchi-etal-2016-personer}.

In another effort to create a standard Persian NER dataset, a corpus was developed by gathering 709 documents from ten different news websites. The authors provided guidelines based on Persian linguistic rules for annotators, resulting in labeled documents as person, location, organization, time, date, percent, currency, or other. The corpus includes 302,530 tokens, with 41,148 tokens labeled as named entities. To ensure inter-annotator agreement, 160 documents were labeled by different annotators, with a Kappa statistic of 95\% \citep{PAYMA}. Additionally, the dataset was used in NSURL-2019 Task 7 \citep{taghizadeh-etal-2019-nsurl}.

\paragraph{English NER Benchmarks}

CoNLL2003 utilized news stories from Reuters between August 1996 and August 1997 for their dataset. They used a segment of 10 days from the end of August 1996 for the training and development set, while the test set was taken from December 1996. For their research, the preprocessed data covered the month of September 1996, which was sourced from the Reuters Corpus2 \citep{conll2033}.

OntoNotes 5.0 is a comprehensive corpus that includes different text genres and languages with syntax and shallow semantic information. It comprises content from prior releases, as well as additional annotations for news, broadcast, telephone conversations, and web data in both English and Chinese. In addition, it contains newswire data in Arabic \citep{Ontonotes}.

The NCBI disease corpus contains 6892 disease mentions that map to 790 distinct disease concepts, with 88\% of them linked to a MeSH identifier and the rest containing an OMIM identifier. While 91\% of the mentions link to a single disease concept, the remaining ones describe a combination of concepts \citep{Dogan2014-il}.

WNUT 2017 released a shared task that focuses on identifying unusual, previously-unseen entities in the context of emerging discussions. Named entities are crucial to many modern approaches to other tasks, such as event clustering and summarization, but annotators often have difficulty in recalling them due to noisy text. This is typically due to novel entities and surface forms. For example, the tweet "so.. kktny in 30 mins?" is challenging for even human experts to detect and resolve the entity "kktny." This task evaluates the ability to detect and classify singleton-named entities in noisy text that are new and emerging \citep{derczynski-etal-2017-results}.

\paragraph{Approaches}

\citet{dandapat2016improved} present a method for improving named entity recognition in Hindi, a resource-poor language. The approach uses cross-lingual information obtained from online machine translation and word alignment. The English named entity recognizer and alignment information are used to estimate cross-lingual features, which are then used in a support vector machine-based classifier. The use of cross-lingual features improves the F1 score by 2.1 points absolute (2.9\% relative) compared to a strong baseline model.

\citet{jain-etal-2019-entity} present a new approach for cross-lingual named entity recognition, which leverages machine translation to improve annotation-projection methods. The system is based on the use of machine translation twice, the matching of entities based on orthographic and phonetic similarity, and the identification of matches based on distributional statistics. The approach shows an improvement of 4.1 points over current state-of-the-art methods and achieves the best F1 score for Armenian, outperforming even a monolingual model trained on Armenian data.
\begin{table}[]
\caption{English (en) benchmarks and generated Persian (fa) datasets number of instances}
\centering
\resizebox{\columnwidth}{!}{%
\begin{tabular}{|c|c|c|c|c|}
\hline
\textbf{dataset}          & \textbf{train} & \textbf{dev} & \textbf{test} & \textbf{avg} \\ \specialrule{.2em}{.05em}{.05em}
CoNLL2003-en              & 14041          & 3250         & 3453          & 15           \\ \hline
CoNLL2003-fa              & 13746          & 3159         & 3380          & 14           \\ \hline
$\Delta$ fa-en                    & -295           & -91          & -73           & -1           \\ \specialrule{.2em}{.05em}{.05em}
OntoNotes 5.0-en & 75187          & 9603         & 9479          & 17           \\ \hline
OntoNotes 5.0-fa & 73907          & 9420         & 9279          & 16           \\ \hline
$\Delta$fa-en                    & -1280          & -183         & -200          & -1           \\ \specialrule{.2em}{.05em}{.05em}
WNUT 2017-en               & 3394           & 1009         & 1287          & 18           \\ \hline
WNUT 2017-fa               & 3386           & 1007         & 1284          & 17           \\ \hline
$\Delta$ fa-en                    & -8             & -2           & -3            & -1           \\ \specialrule{.2em}{.05em}{.05em}
NCBI disease-en          & 5433           & 924          & 941           & 25           \\ \hline
NCBI disease-fa          & 5383           & 916          & 927           & 23           \\ \hline
$\Delta$ fa-en                    & -50            & -8           & -14           & -2           \\ \specialrule{.2em}{.05em}{.05em}
\end{tabular}%
}
\label{table:datasets-info}
\end{table}

\citet{9443126} introduce the Persian Question Answering Dataset (ParSQuAD), which is a translation of the well-known SQuAD 2.0 dataset. The dataset comes in two versions: one that has been manually corrected and one that has been automatically corrected. It provides the first large-scale QA training resource for Persian, a language for which less research has been done in the field of Question Answering due to the lack of datasets. The article reports results from training three baseline models on the dataset, with the best model achieving an F1 score of 70.84\% and an exact match ratio of 67.73\% on the manually corrected version.

\section{DATASET GENERATION}
\label{sec:data}
In the course of developing a Persian benchmark dataset for named entity recognition, we selected four English-language benchmark datasets. Although there are various multilingual pre-trained language models \cite{mohammadshahi-etal-2022-small, chung2022scaling, NLLB} and fine-tuned machine translation models \cite{sartipi-MT} available for Persian-English translation, we simply utilized Google Machine Translation \citep{googlemt}. However, translating sentences without taking into account named entity spans could result in the loss of the order of tokens, and it could also make it challenging to assign named entity tags to each token. To address this challenge, we implemented a strategy for masking named entity spans with a specific format. This format, which is not translated during the machine translation process, preserves the place of the entity in the target language. This approach enables us to maintain consistency in the dataset and correctly tag the named entities in the translated sentences.
Additionally, we recognized that each sentence could contain multiple entities, and the order of these entities in the translated sentence may differ from the order in the source language. To resolve this issue, we added an index to the special format we used for the named entities. This index enables us to retrieve the correct boundaries for each entity in the target language. Thus, we were able to ensure that our Persian benchmark dataset contains accurate and consistent annotations for named entities.
Our automated process of creating a dataset for named entity recognition from a English dataset involves two main steps. In the first step, a given sentence S with N tokens, represented as S = [x1, x2, ..., xn], is accompanied by a corresponding array of named entities T, with N tags, represented as T = [t1, t2, ..., tn]. The named entities in the sentence are extracted, with each entity represented as a single token enclosed within special characters, such as "[*{i}*]". The value of i represents the order of appearance of the named entities within the source text. In the second step of the process, the extracted tokens are joined to form a complete sentence, which is then passed through a machine translation engine, in our case Google Translate. To ensure accurate translation of named entities, phrases or tokens that comprise named entities are also joined and passed to the translation engine separately. The translated named entities are then inserted into the translated sentence based on the previously established alignment of index positions. The overall process of translation for example is demonstrated in Figure \ref{fig:ch1-arch}.

To verify the accuracy of the alignment process, a set of criteria is applied to the translated sentences. Sentences that do not have matching numbers of special patterns in the source and target languages are excluded from the dataset. Additionally, the number of tokens and NER tags for each instance is checked to ensure consistency. A table summarizing the number of instances in the English dataset and the corresponding number of instances in the created dataset is presented in Table \ref{table:datasets-info}. As can be seen in the table, the automated translation process was successful in producing a large number of translated instances. However, a small number of instances could not be translated from the source to the target language, either due to limitations of the machine  translation engine or the inherent complexity of the sentence structure. It is noteworthy that there exist slight differences in the average number of tokens between the target and source languages. 

\section{Experiments}
\label{sec:expr}

Previous research has explored various techniques for developing NER systems \citep{8109733, 8661067, Moradi2017, 7288806, Balouchzahi, jalali-farahani-ghassem-sani-2021-bert, ParsNER}.
To evaluate the quality of the generated datasets, we used the Hugging Face trainer \citep{wolf-etal-2020-transformers} and the xlm-roberta-base \citep{xlm-r} model, which is a multilingual model capable of supporting both English and Persian. We chose this model to ensure that the results were easily interpretable and to allow for direct comparison between the original English datasets and their translated Persian counterparts.
Table \ref{tab:exp-results} presents the results of our evaluation for both the English and translated versions of the main datasets. The rows where the Model column is $\Delta$ tar-src show the difference between the English and Persian datasets. 
Our results indicate that the largest discrepancy between the source and target datasets in terms of F1 score was observed for the WNUT 2017 dataset, while the smallest difference was noted for the NCBI Disease dataset. The former exhibits more complex and sophisticated sentences with emerging named entities that are inherently challenging to recognize and classify, resulting in a relatively lower F1 score for the translated dataset compared to the source dataset. In contrast, the NCBI Disease dataset comprises only three named entity classes, which are relatively easier to recognize and classify, thus demonstrating a smaller difference in the F1 score between the source and target datasets.
In addition to using the multilingual model, we also evaluated a monolingual model called Pars Bert \citep{pars} on the generated datasets. Our experiments revealed that, while the F1 scores for three out of four datasets were slightly lower compared to the multilingual model, the NCBI Disease dataset exhibited a slightly higher F1 score using the Pars-Bert model. These findings suggest that the choice of model may have varying impacts on the performance of the named entity recognition task for different datasets and languages.
\begin{table}[]
\caption{Experiment results on two language models and value of overall F1 (F1), precision (P), and recall (R)}
\centering
\resizebox{\columnwidth}{!}{%
\begin{tabular}{|c|c|c|c|c|}
\hline
\textbf{model} & \textbf{dataset} & \textbf{F1} & \textbf{P} & \textbf{R} \\ \specialrule{.2em}{.05em}{.05em}
xlm-roberta-base            & CoNLL2003-en     & 91.51       & 90.59      & 92.45      \\ \hline
xlm-roberta-base            & CoNLL2003-fa     & 85.11       & 84.49      & 85.76      \\ \hline
Pars-Bert      & CoNLL2003-fa     & 84.04       & 82.07      & 86.11      \\ \hline
$\Delta$ fa-en         & -                & -6.39       & -6.09      & -6.69      \\ \specialrule{.2em}{.05em}{.05em}
xlm-roberta-base            & OntoNotes 5.0-en   & 89.14       & 88.62      & 89.69      \\ \hline
xlm-roberta-base            & OntoNotes 5.0-fa   & 83.95       & 83.96      & 83.94      \\ \hline
Pars-Bert      & OntoNotes 5.0-fa   & 82.80       & 82.8       & 82.82      \\ \hline
$\Delta$ fa-en         & -                & -5.2        & -4.66      & -5.75      \\ \specialrule{.2em}{.05em}{.05em}
xlm-roberta-base            & NCBI disease-en & 83.5        & 83.45      & 83.54      \\ \hline
xlm-roberta-base            & NCBI disease-fa & 83.46       & 81.86      & 85.13      \\ \hline
Pars-Bert      & NCBI disease-fa & 81.71       & 79.52      & 84.02      \\ \hline
$\Delta$ fa-en         & -                & -0.04       & -1.6       & 1.59       \\ \specialrule{.2em}{.05em}{.05em}
xlm-roberta-base            & WNUT 2017-en      & 53.0        & 61.09      & 46.80      \\ \hline
xlm-roberta-base            & WNUT 2017-fa      & 40.02       & 44.98      & 36.05      \\ \hline
Pars-Bert      & WNUT 2017-fa      & 40.31       & 48.43      & 34.52      \\ \hline 
$\Delta$ fa-en         & -                & -12.98      & -16.11     & -10.75     \\ \specialrule{.2em}{.05em}{.05em}
\end{tabular}%
}
\label{tab:exp-results}
\end{table}

\begin{figure}[h]
  \begin{subfigure}[b]{1\columnwidth}
    \includegraphics[width=\linewidth]{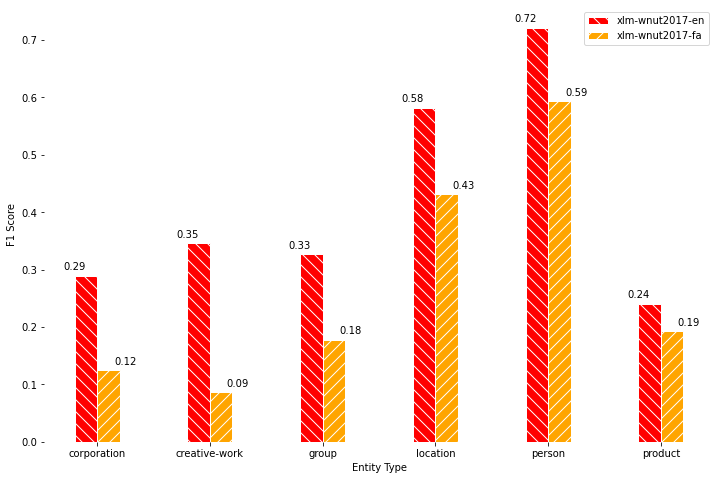}
    \caption{WNUT 2017}
    \label{fig:wnut}
  \end{subfigure}
  \begin{subfigure}[b]{1\columnwidth}
    \includegraphics[width=\linewidth]{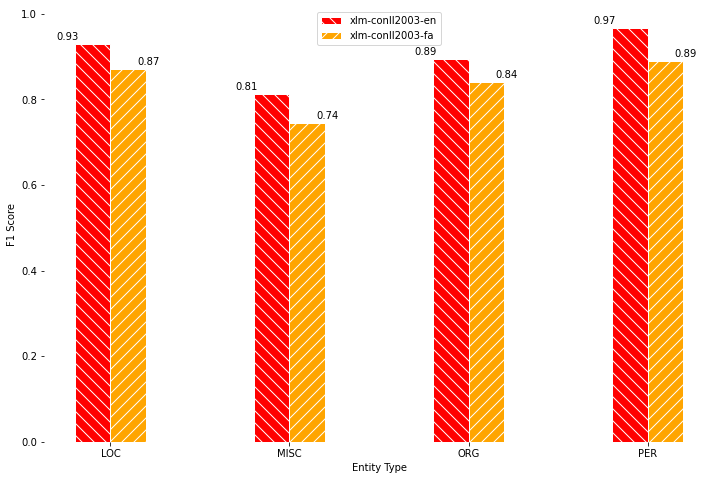}
    \caption{CoNLL 2003}
    \label{fig:CoNLL}
  \end{subfigure}
   \label{fig:wnut-CoNLL}
    \caption{F1 results for WNUT 2017 and CoNLL2003}
\end{figure}

Generated datasets and fine-tuned models on translated English benchmarks are publicly available in Hugging Face\footnote{\url{https://huggingface.co/Amir13}} GitHub\footnote{\url{https://github.com/amirsartipi13/Translated-English-Benchmarks-to-Persian}} repository.



\section{Discussion}
\label{sec:disc}
The CoNLL 2003 dataset comprises four distinct entity classes, namely, Location (LOC), Miscellaneous (MISC), Organization (ORG), and Person (PER). As illustrated in Figure \ref{fig:CoNLL}, the most significant variance between source and target entity types is observed for the Person class, with a difference of 8\%, while the smallest gap is for the Organization class, with a difference of 5\%. Generally, it can be observed that the translated output yields comparable results when contrasted with the source language.

Among the datasets that were experimented with, the OntoNotes 5.0 dataset stands out due to its greater number of classes, with a total of 18 classes. However, it should be noted that in certain classes, such as Event, FAC and WORK\_OF\_ART, as well as LOC, there were significant differences between the English and translated datasets. Specifically, the translations of these classes exhibited significant variations in terms of both precision and recall metrics. These discrepancies could be attributed to a range of factors, including linguistic variations and differences in language structure between the source and target languages. On the other hand, in some classes there were considerably less differences between the English and translated datasets, and in some cases, the results were too close to call. These observations can be seen in the f1 scores for each entity type, which are displayed in Figure \ref{fig:ontov5}.

Because the WNUT 2017 is a dataset for emerging and previously unseen entities, their results not only are lower in the target language, but also they are low in the source language. This could represent that performance of the model on the target side could have a direct correlation between how much source text data are sophisticated and whether their entities are complex or not. As we can see in Figure \ref{fig:wnut}, Location and Person which are more common entities have higher F1 in comparison to creative-works that include names of books or movies.

It is worth noting about the WNUT 2017 that the performance of models on this dataset is generally lower not only in the target language, but also in the source language. This observation suggests that the performance of a given model on the target language has a direct correlation with the sophistication of the source text data and the complexity of the entities involved. 

Notably, the performance of the models on NCBI disease is consistent, with only slight differences observed in the F1 scores between the English and translated versions. We believe that there are two main reasons for this finding. First, it is worth noting that the NCBI disease dataset is a relatively simple dataset, with only three classes that define the beginning (B-disease) and inside token (I-disease) of a disease. This simplicity may have made it easier for the translation model to accurately identify and classify disease entities, resulting in consistent performance across the English and translated versions of the dataset. Second, it should be noted that the machine translation engine used in this study may not have translated the disease names directly, as there may not be exact translations for many of the diseases in Persian. Instead, the translation engine may have simply converted the disease names and written them using Persian characters. 

This approach may have contributed to the consistency of the results, as the Persian characters used to represent the disease names may have been easily recognizable by the named entity recognition model.

Indeed, an analysis of the F1 scores for different entity types reveals interesting trends in the performance of the models. In particular, it can be observed that more common entity types, such as Location and Person, tend to have higher F1 scores compared to more specialized entities like Creative-Works, which include the names of books or movies. This could be attributed to the fact that more common entities are easier to recognize and classify, whereas specialized entities are more challenging to identify and require more context and knowledge to be correctly recognized.

\section{Conclusion}
\label{sec:conc}
In conclusion, our study demonstrates that the simple approach we used to generate a named entity recognition dataset from a high-resource language, specifically English, to a low-resource language, specifically Persian, is highly effective. We found that this approach is particularly useful in the context of specific domains such as biomedical, where the availability of annotated data is limited. Moreover, our approach can be applied to augment existing data and increase the number of instances, which can make the resulting model more robust and reliable.
Our evaluation revealed that certain types of named entities, such as WORK\_OF\_ART and EVENT in the OntoNotes 5.0 dataset, as well as creative-work, corporation, group, and product in the WNUT 2017 dataset, are more challenging to recognize in both English and Persian. These findings highlight the need for continued research into the development of more sophisticated models that can accurately identify and classify such complex named entities in a variety of languages. Overall, our approach offers a promising solution for addressing the challenges of generating high-quality named entity recognition datasets in low-resource languages.

\label{sec:bibtex}
\section*{Acknowledgements}
This work has been supported by the Simorgh Supercomputer - Amirkabir University of Technology under Contract No ISI-DCE-DOD-Cloud-900808-1700.
\nocite{*}

\bibliography{acl_latex}
\bibliographystyle{acl_natbib}





\end{document}